\documentclass[sigconf,authorversion,nonacm]{acmart}
\AtBeginDocument{%
  \providecommand\BibTeX{{%
    \normalfont B\kern-0.5em{\scshape i\kern-0.25em b}\kern-0.8em\TeX}}}

\usepackage{algpseudocode}
\usepackage{algorithm}
\usepackage{amsmath}
\usepackage{dblfloatfix} 
\usepackage{multirow}
\usepackage{multicol}
\begin{document}


\title{Explaining Groups of Instances Counterfactually for XAI: \\ A Use Case, Algorithm and User Study for Group-Counterfactuals}


\author{Greta Warren}
\email{greta.warren@insight-centre.org}
\affiliation{%
  \institution{University College Dublin}
  \streetaddress{Dublin}
  \city{Dublin}
  \country{Ireland}}
\authornote{Both authors contributed equally to this research.}

\author{Christophe Guéret}
\affiliation{%
  \institution{Accenture Labs}
  \streetaddress{}
  \city{Dublin}
  \country{Ireland}}

\author{Mark T. Keane}
\affiliation{Insight Centre for Data Analytics}
\affiliation{VistaMilk SFI Research Centre}
\affiliation{%
  \institution{University College Dublin}
  \streetaddress{Dublin}
  \city{Dublin}
  \country{Ireland}}

\author{Eoin Delaney}
\email{eoin.delaney@insight-centre.org}
\authornotemark[1]
\affiliation{VistaMilk SFI Research Centre}
\affiliation{%
  \institution{University College Dublin}
  \streetaddress{Dublin}
  \city{Dublin}
  \country{Ireland}}


\begin{abstract}
Counterfactual explanations are an increasingly popular form of post hoc explanation due to their (i) applicability across problem domains, (ii) proposed legal compliance (e.g., with GDPR), and (iii) reliance on the contrastive nature of human explanation. Although counterfactual explanations are normally used to explain individual predictive-instances, we explore a novel use case in which  groups of similar instances are explained in a collective fashion using ``group counterfactuals'' (e.g., to highlight a repeating pattern of illness in a group of patients). These group counterfactuals meet a human preference for coherent, broad explanations covering multiple events/instances. A novel, group-counterfactual algorithm is proposed to generate high-coverage explanations that are faithful to the to-be-explained model. This explanation strategy is also evaluated in a large, controlled user study (N=207), using objective (i.e., accuracy) and subjective (i.e., confidence, explanation satisfaction, and trust) psychological measures. The results show that group counterfactuals elicit modest but definite improvements in people's understanding of an AI system. The implications of these findings for counterfactual methods and for XAI are discussed.
\end{abstract}


\keywords{XAI, Counterfactual, Group Explanation, User Study}

\maketitle              
\section{Introduction}
In recent years, the literature on eXplainable AI (XAI) has focused significantly on the use of counterfactual explanations to explain the predictions of opaque machine learning (ML) models \cite{miller2019explanation,goyal2019counterfactual,guidotti2018survey,karimi2021algorithmiccf,keane2020good,kenny2021generating}, as they can show what changes to input-features can alter a model's output decisions (e.g., "if only the bank customer had applied for a lower loan amount of \$10,000, their loan application would have been approved"). Interest in counterfactuals as explanations has been boosted by arguments made in philosophy and psychology about the formal analysis of causality \cite{lewis2013counterfactuals,woodward2005making} and people's causal thinking \cite{Byrne2019CounterfactualsReasoning,miller2019explanation}, respectively. Furthermore, legal analyses have suggested that counterfactual explanations comply with GDPR requirements \cite{wachter2017counterfactual}, leading to their extensive use in algorithmic recourse \cite{karimi2020algorithmic}. Indeed, this XAI literature now comprises over $\sim$120 counterfactual methods, all of which claim to generate plausible counterfactual explanations that people understand \cite{keane2021if,verma2022counterfactual}. However, these claims of psychological validity have not been supported by appropriate user tests; a 2021 review \cite{keane2021if} reports that only $\sim$7\% of counterfactual methods had been directly user-tested. 

Counterfactual XAI has also been criticised for not paying sufficient attention to identifying appropriate use cases, failing to determine when counterfactual explanations work well and when they might fail \cite{arya2021one,sokol2020one,barocas2020hidden}. Indeed, the mixed findings emerging from recent user studies may actually arise from inappropriate applications of counterfactual explanations (see e.g., \cite{Warren2022FeaturesXAI,Warren2023Features2,lage2019evaluation,celar_cfactual23,van2021evaluating,Dai2022CounterfactualXAI,kuhl2022keep,kuhl2022let}). In this respect, it is perhaps significant that the current work arose from end-user feedback collected during focus groups trialling AI models (specifically, dairy farmers testing animal disease prediction models). Most counterfactual-explanation use cases assume a one-shot scenario, in which a prediction is made for a single instance, and the user is provided with a single explanation (e.g., the classic lending case \cite{grath2018interpretable}). In contrast, recent end-user evaluations have suggested use cases in which predictions are made for multiple similar instances with the same outcome, prompting the novel strategy of grouping these multiple predictive-instances (implicitly), by using the ``same'' counterfactual explanation (i.e., feature-differences using an identical target-value, see Figure \ref{fig:fig1}). For example, if a farmer has several sick animals with similar characteristics that can be explained by the same counterfactual then, intuitively, it becomes easier to spot disease patterns, and also perform more consistent recourse in the treatment of the disease.

\subsection{When Explanations for Groups Might Help}
\label{when_group_exp}
Recent research in the field of Smart Agriculture (SmartAg) has attempted to use ML techniques to support more sustainable and ethical agriculture by improving disease prediction and supporting better agricultural practices that prioritise animal health and welfare (e.g., by avoiding the overuse of antibiotics). For example, \textit{mastitis} -- an inflammation of a milking cow's udder -- is a disease which significantly impacts animal health and welfare in the dairy sector. As the onset of mastitis is quite hard to predict, inappropriate treatments can be administered, in which both sick and healthy animals are given antibiotics (leading to antibiotic resistance in herds, with many consequential negative effects). Accordingly, recent SmartAg research aims to better predict the onset of this disease \cite{arjun2023predicting} and to explain these predictions sensibly to farmer end-users \cite{ryan2021predicting}. Specifically, these AI models predict when an individual cow is likely to fall ill, so they can be isolated from the herd to prevent contagion and receive appropriate treatment. To explain these predictions, Ryan et al. \cite{ryan2021predicting} used counterfactual XAI techniques. For instance, an animal with a high white blood-cell count ($>$100 units) might be predicted to fall ill in the coming week, thus a counterfactual explanation for this prediction may inform the farmer: ``if this animal had a lower white blood-cell count (e.g., $<$100 units), it would have been predicted to be healthy in the coming week".

Reported work with focus groups in this application area \cite{ryan2021predicting} revealed that farmers were likely to see predictions for several similar animals with common features (e.g., five cows with elevated cell-counts of $\sim$100 units, all predicted to fall ill). In such scenarios, using a common explanation for similar predictive-instances makes sense. Instead of generating five different counterfactuals, one for each animal (e.g., ``if cow-1 had a lower cell-count of 19 then it would not fall ill'', ``if cow-2 had a lower cell-count of 74 then it would not fall ill'' and so on), it seems better to find common feature-differences that explain all five animals (e.g., ``if cow-1 had a lower cell-count of \textit{54} then it would not fall ill'', ``if cow-2 had a lower cell-count of \textit{54} then it would not fall ill'' and so on). This \textit{group counterfactual} strategy potentially provides a consistent counterfactual explanation (implicitly) signalling that these instances are a related group. Farmer focus groups maintained that these explanations would be much more informative about patterns of illness in the herd and better inform herd-level decisions (e.g., about disease treatment). Presenting users with these related counterfactuals for several predictive-instances rather than different counterfactuals for each instance may also have psychological advantages, as it should reduce memory load and facilitate pattern-finding \cite{keane2020good}.

Finally, this scenario is clearly not unique to SmartAg. For example, in medicine, related use cases occur in the diagnosis and treatment of human illnesses. In Smart Manufacturing, related scenarios occur in the prediction and explanation of breakdowns in identical machines on a production line. Indeed, the use of group counterfactuals should apply to any situation in which users encounter multiple predictions over instances that bear similarities in input features and outcomes, and where the user needs to spot patterns over these instances reflecting underlying regularities in the domain. In the next section, we elaborate on how this idea could be applied in an income-prediction domain, to work up this proposal from scratch, and to sketch how group counterfactuals might be computed (see Figures \ref{fig:fig1} and \ref{fig:fig2}).

\subsection{Explaining Multiple Income Predictions} 

\begin{figure*}[!h]
  \centering
  \includegraphics[width=\textwidth]{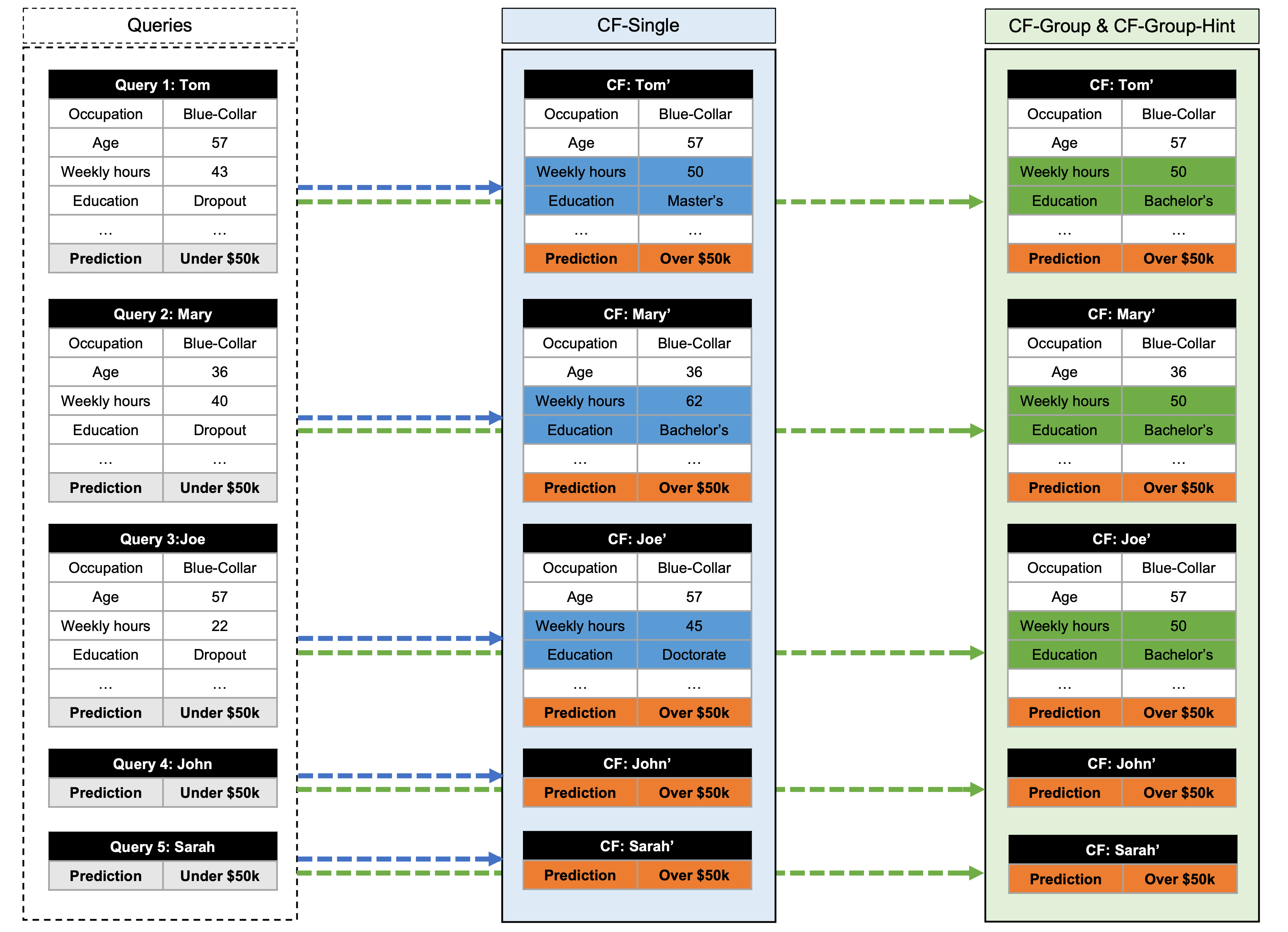} 
  \caption{Sample Queries with Single or Group Counterfactual Explanations.  Five queries are shown in summary form  with matched single or group counterfactual explanations (n.b., details for John and Sarah omitted).  Note that target-values for Weekly Hours and Education in the single counterfactuals vary for each explanation, whereas those for the group counterfactuals are the same in each. By definition, all these counterfactual feature-differences flip the outcome class.}
  \label{fig:fig1}
\end{figure*}

In the current paper, we used the Adult\footnote{The New Adult Datasets \cite{ding2021retiring} could be used here and may be preferred if examining group counterfactual explanations through the lens of fairness in XAI (see also \cite{kasirzadeh2021use}).} (Census) dataset \cite{Dua2019UCI} to build a classifier that predicted whether individuals earned under or over \$50,000 in annual income (see Figure \ref{fig:fig1}). In our envisioned use case, a certain level of domain expertise can be assumed for the user evaluation of explanations (e.g., farmers receiving mastitis predictions), thus our motivation for using the Adult dataset was to approximate this when evaluating it with participants from a general population. We developed a novel method -- called \textbf{Group-CF} -- that explains the classifications of similar individuals in the same class using a group counterfactual, implicitly collecting them together using common feature-differences (see also Figure \ref{fig:fig1}). In this use case, we assume an administrator is checking the classification of multiple individuals (e.g., a bank clerk assessing customer risk). To evaluate this new explanation strategy with users, two different counterfactual XAI methods were implemented to produce test materials, the: 

\begin{itemize}
 \item \textit{CF-Single} system that explained classifications using diverse, single counterfactuals for each predictive-instance based on diverse feature-differences. For instance, ``if Tom had worked \textit{50} hours per week instead of 43, then he would have earned over \$50k'', ``if Mary had worked \textit{62} hours per week instead of 40, then she would have earned over \$50k'', ``if Joe had worked \textit{45} hours per week instead of 22, then he would have earned over \$50k'' and so on (see also example in Figure \ref{fig:fig1}). 

\item  \textit{CF-Group} system, using the Group-CF method, that explained classifications using a group counterfactual covering several predictive-instances, re-using the same target-value in the feature-differences. For instance, ``if Tom had worked \textit{50} hours per week instead of 43, then he would have earned over \$50k'', ``if Mary had worked \textit{50} hours per week instead of 40, then she would have earned over \$50k'', ``if Joe had worked \textit{50} hours per week instead of 22, then he would have earned over \$50k'' and so on (see also example in Figure \ref{fig:fig1}). 

\end{itemize}

\begin{figure*}[!h]
  \centering
  \includegraphics[scale=0.5]{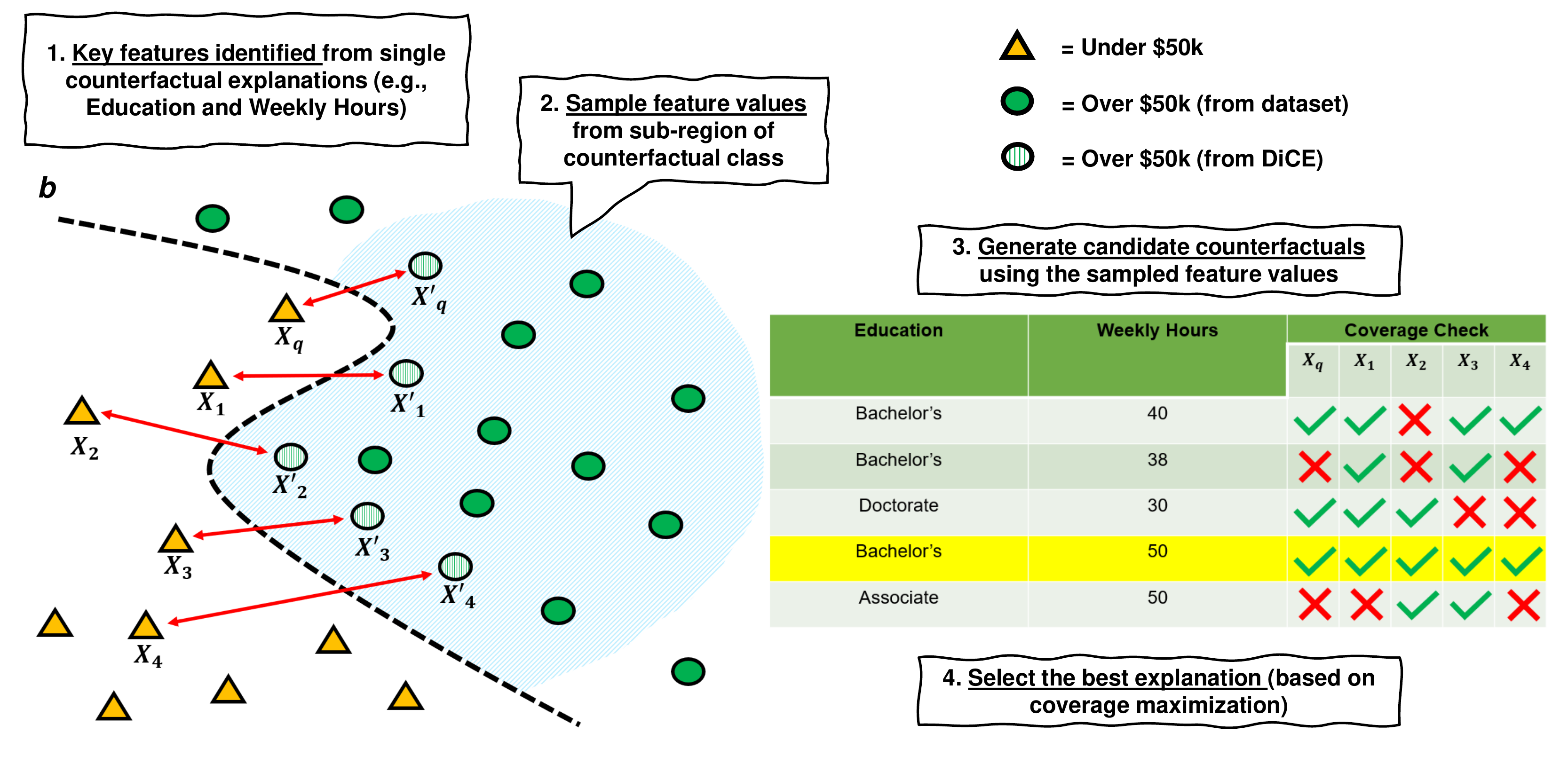}
  \caption{
  A related group of queries,  $(X_{q}, X_{1}, X_{2}, ..., X_{n})$, are classed as being on one side of a decision boundary (e.g., Under \$50k). Single counterfactuals are generated for each of these points forming a set of individual explanations, to find the feature-differences most commonly used (i.e., Education and Weekly Hours). Values for these identified features are sampled from a region of training instances in the contrasting class (i.e., Over \$50k). These values are substituted into the query-instances to create candidate group counterfactuals that are checked for validity, with the best being chosen based on coverage.}
  \label{fig:fig2}
\end{figure*}

\noindent Figure \ref{fig:fig1} shows a detailed two-feature-difference, worked example of how these two systems differ in the counterfactual explanations they generate for the same predictive-instances. Figure \ref{fig:fig2} graphically represents the main steps in the Group-CF  algorithm used in the CF-Group system. In a user study, participants in the CF-Single condition were presented with classifications that were explained using diverse, single counterfactual explanations (i.e., from the CF-Single system). Participants in the CF-Group and CF-Group-Hint conditions were presented with the same classifications, which were explained using group counterfactuals for related classified instances (i.e., from the CF-Group system). The participants in the CF-Group-Hint condition received an additional ``hint'' along with each explanation that informed users that the individual belonged to a related group of people, to explicitly signal the commonality between the counterfactual instances.  


We hypothesised that because these group counterfactuals present users with the same values in key feature-differences, they represent a broader and more coherent explanation \cite{lombrozo2006structure,lombrozo2016explanatory,thagard1989extending,edwards2019explanation} than single counterfactuals, which vary in the feature-changes they suggest. Thus, group counterfactual explanations should facilitate a better understanding of the AI system and domain (as evidenced by higher user prediction accuracy). We also predicted that this improved understanding engendered by group counterfactuals would lead to greater user confidence in their own predictions, as well as elicit higher satisfaction and trust in the AI system, relative to single counterfactuals.

\subsection{Outline of Paper \& Contributions}

The remaining sections of the paper provide more detail on the proposed group-counterfactual algorithm (section \ref{algorithm}), the methodology for the user study (section \ref{user_study}) and its results (section \ref{user_study_results}). Though there is very little work on this type of counterfactual explanation in both AI and Psychology, we try to identify the most relevant related work (in section \ref{related_work}), before concluding with caveats, limitations and future directions (section \ref{conclusions}). The paper makes several novel contributions in use case definition, algorithmic development, and user testing:

\begin{enumerate}
 \item \textit{Use Case Definition}: a first formulation of the \textit{multi-prediction use case} in counterfactual XAI, a scenario in which end-users encounter multiple, similar predictive-instances generated by a single AI system that are explained by a common explanation revealing related occurrences and patterns in predictive outcomes
 
\item \textit{Algorithmic Development}: a novel counterfactual XAI algorithm -- the \textbf{Group-CF} method -- that groups explanations of similar predictive-instances, along with a methodology for presenting these to end-users

\item \textit{User Testing}: the first user evaluation of this new group counterfactual XAI method, carefully designed and executed to reveal the key impacts it has on objective (i.e., accuracy) and subjective (i.e., confidence, satisfaction and trust) psychological measures of human understanding

\end{enumerate}

\begin{algorithm}[!h]
\caption{Group-Counterfactual Method}
\label{alg:cap}
\begin{algorithmic}
\Require $b(.);$ to-be-explained black-box model
\Require $D_{c}$; instances in the  training data with class label $c$
\Require $D_{c'}$; instances in the  training data with class label $c'$
\Require $X_{q};$ Query instance described by a vector of features $[f_{1}, f_{2}, ..., f_{k}]$ such that $b(X_{q})$ = $c$

\State Retrieve, $X$ $\in$ $D_{c}$, the Nearest Like Neighbor subset pool for Query $[X_{1}, X_{2}, ..., X_{n}]$ $\in$ $X$  are selected such that $b(X)$ = $c$.

\For{$x \in \{X_{q}\}\bigcup\{X_i\}_{i=1}^{n}$}
    \State Generate individual CFE, $X'$, s.t. $b(X')$ = $c'$. 
    \State Note the feature change and the direction if applicable.
\EndFor    
\State The most commonly perturbed feature set from the individual counterfactuals informs the features changed in the Group-CF.
\State Randomly sample feature pairs $[f_{i}, f_{j}]$ from sub-region $R$ $\in$ $D_{c'}$ 

\For{$[f_{i}, f_{j}]$ in samples}

    \For{$X \in \{X_{q}\}\bigcup\{X_i\}_{i=1}^{n}$}
        \State substitute feature values with $[f_{i}, f_{j}]$ $\gets X_{sub}$
        \State If $b(X_{sub}) \neq c'$
    \EndFor
\State Return sample feature pair that maximises coverage 
\EndFor
\State \textbf{Stop}
\\
\\





\end{algorithmic}
\end{algorithm}

\section{Group-Counterfactual Algorithm} 
\label{algorithm}

Figure \ref{fig:fig2} graphically represents the main steps in the \textbf{Group-CF} algorithm developed to meet the multi-explanation use case defined earlier (see Algorithm 1 for formalization). The algorithm's starting point is a set of related instances that have been classified by the model to be from the same class; for instance, a pool containing $X_{q}$ and its nearest-like-neighbours $(X_{1}, X_{2}, .... X_{n})$ (n.b., pool size is a hyperparameter, here set to 5). Taking these inputs the four main steps of the method (i) \textit{identifies key difference-features} by generating individual counterfactuals for each related instance and analysing the feature differences on which they rely, and then (ii) \textit{samples feature-values} for key difference-features from data-points in the contrasting class to (iii) \textit{generates candidate group-counterfactuals} based on substituting these feature-values into the original instances before (iv) \textit{selects the best group-counterfactual} based on its valid coverage of all the instances in the pool.  Note, this method does \textit{not} attempt to generalise the individual counterfactuals for the pool of instances, though it does use these counterfactuals as a guide to the features that might be useful in good group-counterfactuals (see section \ref{sec:materials} for implementation details and GitHub link for code - \url{https://github.com/e-delaney/group_cfe})


\begin{figure*}[!b]
\centering
\includegraphics[width=0.85\textwidth]{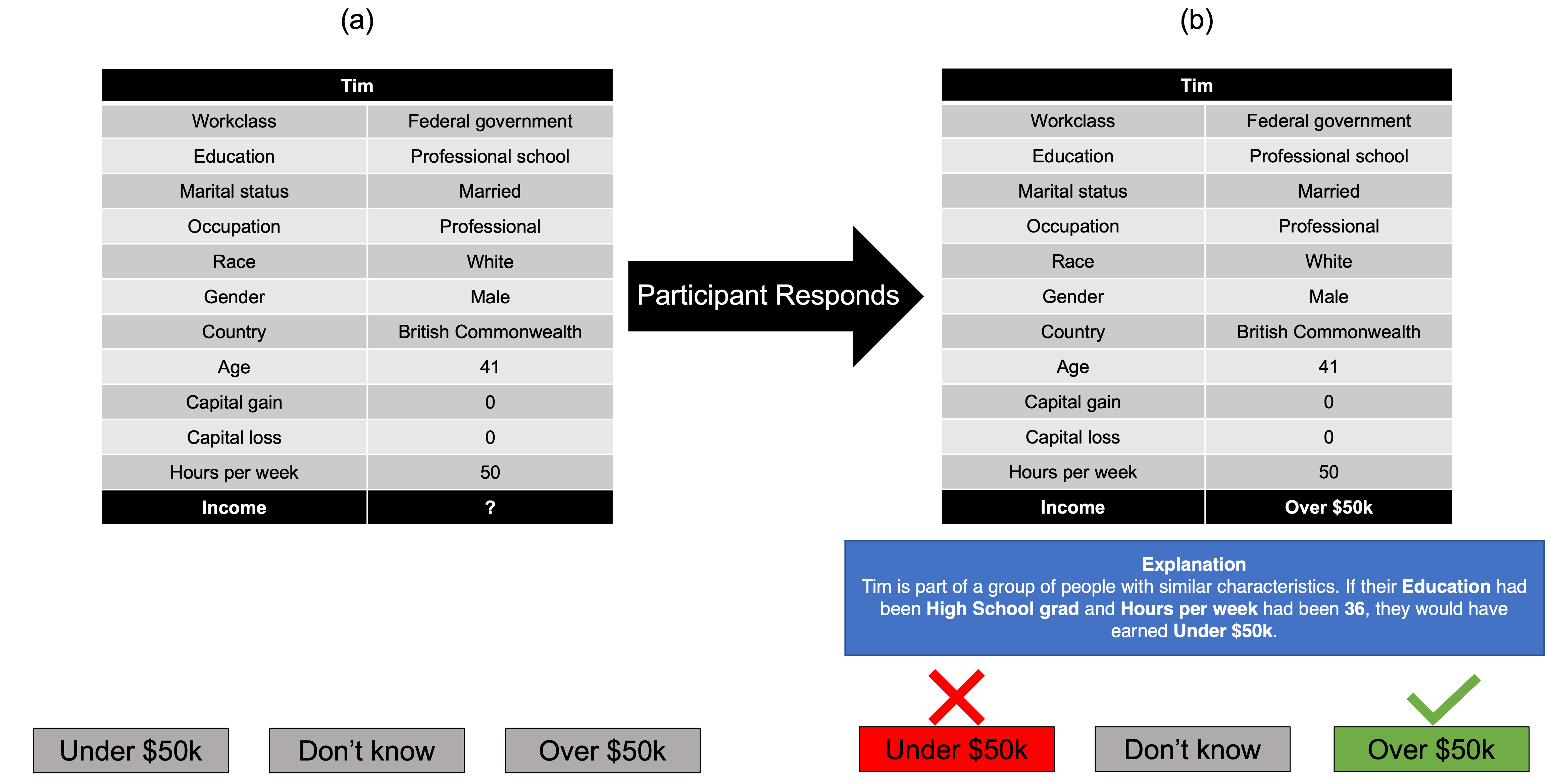}
\Description{Multipart figure showing the interface of the income prediction app during the training phase with feedback specific to the explicit group counterfactual explanation group. (a) shows a two-column table displaying the tabular information for an instance, and below it are three grey response buttons side by side and labelled 'Over \$50k', 'Don't know' and 'Under \$50k' respectively. (b) shows the same instance, and now underneath is a text box with an explanation reading “Tim is part of a group of people with similar characteristics. If their Education had been High School grad and Hours per week had been 36, they would have earned Under \$50k” At the bottom of the screen are the three options in separate text boxes: 'Over \$50k', 'Don't know' and 'Under \$50k'. The “Over \$50k” box is coloured green, with a green tick mark over it, and the “Under \$50k” button is red instead, with a red ‘X’ mark over it.}
\centering
\caption{Sample Material from Training Phase of the user study. The participant first sees (a) with the task of providing their own prediction for the item (3 options). After they respond they are presented with (b), showing feedback on the correctness of their response along with an explanation (in this case for the CF-Group-Hint condition).}
\label{fig:fig3}
\end{figure*}

\subsection{Step 1: Key-Feature Identification}
Given a pool of to-be-explained instances -- $X_{q}$ and its nearest-like-neighbours  $(X_{1}, X_{2}, .... X_{n})$ -- and an opaque black-box model, $b$, ensure that\footnote{It is not enough just to select training instances with class label $c$ as the model may not agree with the class label due to regularization effects of the classifier.} $b(X)$ = $c$. Then, generate individual counterfactuals for all instances in the pool, forming a set of counterfactual instances $(X'_{q}, X'_{1}, X'_{2}, ..., X'_{n})$ such that $b(X')$ = $c'$. Any ``traditional'' counterfactual method can be used to generate these diverse counterfactuals; we used DiCE \cite{mothilal2020explaining} due to its popularity, performance beyond baseline methods, and the availability of well-maintained open source code. Then, the difference-features on which these counterfactuals rely are analysed, to find the most commonly-used features that produce contrasting instances.  For example, Education and Weekly Hours emerge here as the most frequently used difference-feature that flips the classification (see Figure \ref{fig:fig1}), so these would be chosen as the key features to modify in generating candidate group-counterfactuals (n.b., majority voting is just one way to do this). This step also identifies the direction-of-difference in the key features used to flip the classification (e.g., increase/decrease for a continuous feature), a property used during the next step. 


\subsection{Step 2: Sample Feature Values} Using the key features that were found to be the most important in single counterfactuals, sample data points in the \textbf{contrasting class} to find feature-values for use in candidate group-counterfactuals. For example, for the Education feature this sampling identifies candidate values such as Bachelor's, Doctorate or Associate's degree (see Figure \ref{fig:fig2}). Importantly, these sampled instances are known to be valid data points, and therefore are more likely to yield feature-value transformations that result in valid counterfactuals (n.b., unlike those generated in Step 1's single counterfactuals which could be synthetic, invalid data points \cite{keane2020good}). The data points sampled are drawn from a region, that could be all the instances in the contrasting class (e.g., the Over \$50k class).  However, it makes more sense to reduce the size of this region (e.g.,  only consider data points with feature-values in the same direction-of-difference as those found earlier).  We have also found that using feature-values from prototype-like instances in the selected region works well (e.g., medoids from $k$-medoids clustering).

\subsection{Step 3: Generating Candidates} 
The feature-values the from region's data points in the contrasting class are substituted for the values in key-features of the original instances to generate candidate group-counterfactuals (see Figure \ref{fig:fig2}). These perturbations are counterfactual transformations of the original instances, that could cover the pool of instances.

\subsection{Step 4: Selecting the Best Explanation}
The feature-value substitutions that create the candidate group-counterfactuals are checked for validity and coverage. The validity check determines whether the feature-changes flip the classification of a given instance to the contrasting class. The coverage check determines whether this classification change holds over all the original instances in the pool. Accordingly, the  group counterfactual with the highest coverage is chosen as the best, to be used to explain multiple instances (see Figure \ref{fig:fig2}).

\section{User study: Method}
\label{user_study}
\begin{figure}[t!]
\centering
\includegraphics[width=0.45\textwidth]{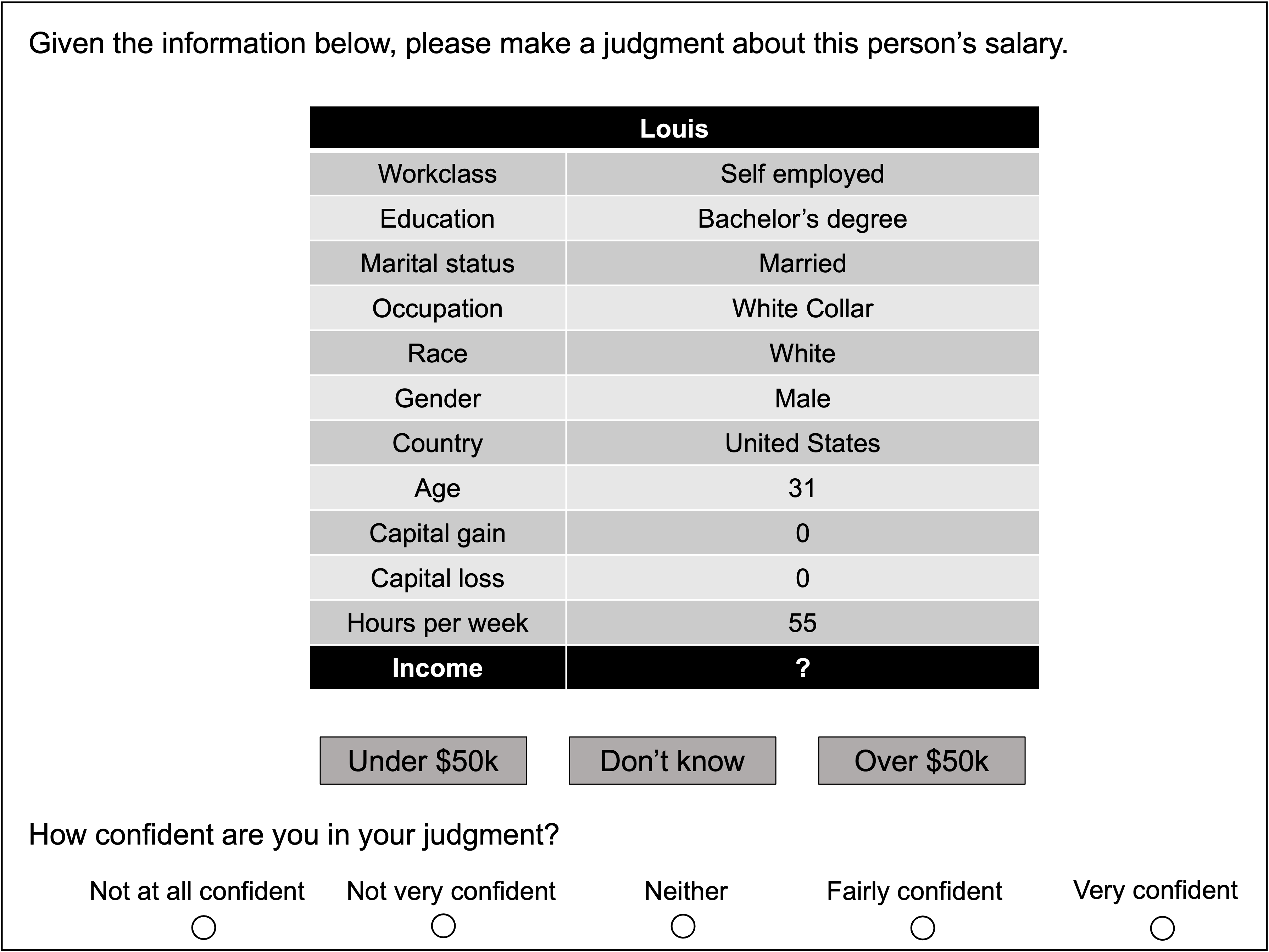}
\Description{Image of app interface during the testing phase of the study. At the top of the screen is text that reads “Please make a judgment about this person's salary” Underneath is a two-column table displaying the tabular information for an instance. At the bottom of the screen are three options in separate grey text boxes: “Over \$50k”, “Don’t know”, “Under \$50k”.}
\centering
\caption{Sample Material in Testing Phase of the User Study}
\label{fig:fig4}
\end{figure}

We designed a user study to determine whether providing counterfactual explanations for a group of predictions improved people's understanding of an AI decision-making system. The study had two phases:a (i) \textit{training phase}, in which people were shown instances from a dataset and asked to predict their outcomes before being shown the AI system's prediction along with an explanation (see Figure \ref{fig:fig3}), and a (ii) \textit{testing phase}, in which people were shown instances and asked to predict their outcomes with no feedback or explanation as to their correctness (see Figure \ref{fig:fig4}). The main measure was \textit{accuracy}, the proportion of instance-items correctly predicted in the testing phase (subjective measures of confidence, satisfaction and trust were also recorded). This train-and-test methodology has previously been used to explore the effects of counterfactual explanations on user understanding in the XAI literature \cite{Dai2022CounterfactualXAI,celar_cfactual23,Warren2022FeaturesXAI,Warren2023Features2,van2021evaluating}. Participants were shown 40 instances in each phase of the study (i.e., 80 unique items in total), with no overlap between the items in each phase. In the training phase, the 40 items were made up of eight 5-item groups of similar instances for which group counterfactuals were generated (in the relevant conditions). Participants were placed into one of three conditions -- CF-Single, CF-Group, or CF-Group-Hint -- that were matched in every respect except for the type of counterfactual explanations provided during the training phase (see section \ref{sec:design} for details).

The main predictions made for the study were that: (i) the provision of explanations (any explanation) will improve accuracy, that is, participant task accuracy in the testing phase will be higher than accuracy in the training phase, (ii) task accuracy in the group-counterfactual conditions will be higher than the corresponding single-counterfactual condition, (iii) the provision of a hint in the CF-Group-Hint condition should improve accuracy relative to the CF-Group condition, (iv) that group-counterfactual explanations should improve confidence in answers, explanation satisfaction and trust in the AI system relative to the single-counterfactual condition.

\subsection{Design}
\label{sec:design}
The study had a 3 (Explanation: CF-Single vs CF-Group vs CF-Group-Hint) x 2 (Phase: Training vs Testing) mixed design with Explanation as a between-participant variable and Phase as a within-participant variable. The three Explanation conditions varied the counterfactual explanations provided in the training phase: \textit{CF-Single} presented people with instance-predictions that were explained using a single counterfactual unique to that instance (e.g., "If Joe's \textit{Weekly hours} had been \textit{45} and his \textit{Education level} had been \textit{Doctorate degree}, he would have earned Over \$50k"; see Figure \ref{fig:fig1}), \textit{CF-Group} presented people with instance-predictions that were explained using group-counterfactuals based on common target-values, implicitly grouping the similar instances in a given 5-item set (e.g., "If Joe's \textit{Weekly hours} had been \textit{50} and his \textit{Education level} had been \textit{Bachelor's degree}, he would have earned Over \$50k"; see Figure \ref{fig:fig1}), and \textit{CF-Group-Hint} presented people with the same group-counterfactuals as in CF-Group along with a ``hint" saying the instance was ``part of a group of people with similar characteristics" (see Figure \ref{fig:fig3}). Note, CF-Single is essentially the control condition, with CF-Group and CF-Group-Hint being the experimental conditions, as it is matched in every respect with the two latter conditions, except in its use of non-group-counterfactual explanations (i.e., single counterfactuals).  

\subsection{Participants}
\label{sec:participants}
Participants (N=207) were recruited using \textit{Prolific Academic} (\url{https://www.prolific.co/}), and assigned in a fixed order to three between-participant conditions: CF-Single (n=68), CF-Group (n=70) and CF-Group-Hint (n=69). The sample consisted of 122 women, 84 men and one non-binary person aged 18-76 years (\textit{M}=40.86, \textit{SD}=14.31), with respondents pre-screened to select native English speakers from Ireland, the United Kingdom, the United States, Australia, Canada and New Zealand, who had not participated in previous related studies. Prior to analysis, 19 participant responses were removed as they failed more than one attention or memory check. An \textit{a priori} power analysis with G*Power \cite{faul2009statistical} indicated that 207 participants were required to achieve 90\% power for a medium effect size with alpha <.05 for two-tailed tests.

\subsection{Materials: Dataset \& Implementations}
\label{sec:materials}
The Adult (Census) dataset describes census information \cite{Dua2019UCI}, and contains a mixture of continuous (age, weekly work hours, capital gain, capital loss), and categorical variables (employment type, occupation, marital status, country of birth, gender, race). A model that predicted a binary outcome, whether a person is earning above or below \$50k/year, was implemented using a Gradient Boosting Classifier \cite{friedman2001greedy}, as the to-be-explained black-box model. 
The default sklearn hyperparameters are used with a log loss implemented and a learning rate of 0.1 achieving an accuracy of 0.874 on the task. Training and test data were split and pre-processed according to \cite{klaise2020alibi} where categorical features were encoded ordinally when possible. Instances were randomly selected from the dataset as query-items for the study. The instances selected were all correct classifications, with low predicted class confidence (within 0.15 of the other class) to make the classifications non-obvious to participants, and the selection was balanced with equal numbers for each class. For the training phase, 8 seed-instances were randomly selected (balanced across classes) and 4 nearest-like-neighbours were found for these seeds (using Hamming distance) to create the 40 queries used (i.e., eight 5-item sets of related instances). For the testing phase, all 40 items were randomly selected from the dataset as the queries to be used, balanced across classes. DiCE \cite{mothilal2020explaining} was used to generate the individual counterfactual explanations. The DiCE-counterfactual explanations (i.e., singles) were generated for each of the 40 training phase queries and used in the CF-Single condition\footnote{We used the original random sampling variant of DiCE implemented using an sklearn backend and a sample size of 1000. The default posthoc sparsity parameter of 0.1 and stopping threshold of 0.5 were implemented.}. For the CF-Group conditions the Group-Counterfactual method (see section \ref{algorithm}), was applied to these the 5-item-instance-sets to find good group-counterfactuals to cover them. If a group-counterfactual was found then the five instances with their paired individual- and group-counterfactuals were used as an item-set in the study. Finally, the sets of counterfactuals used in the CF-Single and CF-Group conditions were matched on proximity and sparsity measured in the standard way \cite{mothilal2020explaining, wachter2017counterfactual}. Proximity was measured using $\ell_{1}$ distance scores, scaled by the median absolute deviation (MAD) of the feature's values in the training set (for continuous features), and a metric that assigns a distance of 1 if the features differ from the original input or zero otherwise (for categorical features). These distance scores were computed for the matched query-explanation pairs used in the control and experimental conditions; a paired, two-tailed t-test showed that they were not significantly different to one another, $t(39)=1.30, p=.197$. Sparsity was measured as the number of feature-differences between counterfactual pairs which was always 2 for all query-explanation pairs used.

\subsection{Measures: Accuracy \& Subjective Measures}
\label{sec:measures}
A mix of objective and subjective measures was used (see \cite{Warren2023Features2} for a discussion of this distinction). The key objective measure was \textit{accuracy}, which measured the extent to which exposure to the model's predictions and explanations  in the training phase improved their knowledge of the domain/model; specifically, it was measured as the proportion of correct responses made in the training phase (i.e., consistent with the model's predictions). The subjective measures evaluated people's self-assessments of their (i) \textit{confidence} in their own predictions made in the testing phase (ii) \textit{satisfaction} with explanations used overall by the AI system, (iii) \textit{trust} in the overall AI system. The latter two measures were made after people completed the training and testing phases of the study using the DARPA Explanation Satisfaction and Trust scales \cite{hoffman2018metrics}. To ensure that participants were fully engaged with the task, they were shown four attention checks at randomised intervals and asked to recall and select a subset of the features used by the system from a list of 10 options (5 correct, 5 incorrect) at the end of the experiment. 

\subsection{Procedure}
\label{sec:procedure}
Ethics approval for the study was granted in advance by University College Dublin with the reference code \textit{LS-E-20-11-Warren-Keane}. At the beginning of the study, participants were informed that they would be testing an AI system designed to predict people's annual income from available information about them. Participants read detailed instructions about the task and provided informed consent. After completing practice trials for each phase of the study, participants progressed through both phases of the main task, and subsequently completed the subjective measures (i.e., Explanation Satisfaction and Trust scales \cite{hoffman2018metrics}). During the training phase, on each screen participants were presented with an instance without the class prediction and asked to make an \textit{income} prediction from three options -- Under \$50k / Don't know / Over \$50k -- as shown in Figure \ref{fig:fig3}(a). Button order was randomised for each item, to prevent users from repeatedly selecting the same response. After making their prediction, users were shown feedback, with the AI model's prediction (correct answer) shown in green with a tick-mark, and the incorrect answer shown in red with a cross-mark. Figure \ref{fig:fig3}(b) shows an example from the Group-CF-Hint condition where a correct prediction was made and explained using a group-counterfactual with a hint. After completing the 40 items in the training phase, people progressed to the testing phase, in which they were shown 40 further instances without outcomes shown (see Figure~\ref{fig:fig4}) and asked to choose one of the three options to make their prediction. Here, after each response, participants rated their confidence in their prediction (using a 5-point Likert scale, from 1 (\textit{Not at all confident}) to 5 (\textit{Extremely confident}). In the testing phase, participants progressed through the 40 items, providing their predictions and confidence judgments without receiving feedback or explanations. After the testing phase, participants completed the satisfaction and trust measures before concluding the study. All the items presented in both phases were randomly re-ordered for each participant to control for possible order effects in the instances seen. On completion, participants were debriefed on the background and motivation for the study and paid £2.50 for their time. The task instructions and data for the study are available at \url{https://osf.io/smupq/?view_only=3ce9d1bafcb74fada825a384a7d24687}.

\begin{table*}[!t]
\centering
\setlength{\tabcolsep}{12pt} 
\begin{tabular}{l cc cc cc}
    \toprule
\multirow{2}{4em}{Measure}
        & \multicolumn{2}{c}{CF-Single} & \multicolumn{2}{c}{CF-Group} & \multicolumn{2}{c}{CF-Group-Hint}   \\
    \cmidrule(lr){2-3} \cmidrule(lr){4-5} \cmidrule(lr){6-7}
        & \textit{M}  & \textit{SD}  & \textit{M}  & \textit{SD}  & \textit{M}  & \textit{SD}  \\
    \midrule
    Accuracy & 0.816 & 0.104 & 0.839 & 0.085 & 0.832 & 0.100 \\
      Confidence   & 3.956 & 0.432    & 3.936 & 0.437    & 4.047 & 0.432    \\
     Explanation Satisfaction   & 26.632 & 6.462    & 25.800 & 6.826    & 28.493 & 6.910    \\
       Trust   & 24.324 & 5.454    & 24.786 & 5.522    & 25.899 & 5.480    \\
    \addlinespace
    \bottomrule
\end{tabular}
\caption{Means and standard deviations for each measure in the conditions of the user study (CF-Single, CF-Group, CF-Group-Hint) for (i) \textit{Accuracy} (proportion of correct answers in the testing phase), (ii) \textit{Confidence} (ratings on a 5-point scale of each answered item in the testing phase), (iii) \textit{Explanation Satisfaction} (summed ratings from the 8-question DARPA-scale after the testing phase), (iv) \textit{Trust} (summed ratings from the 8-question DARPA-scale after the testing phase).}
\label{tab:mean_scores}
    \end{table*} 

\section{User study: Results \& Discussion}
\label{user_study_results}
Overall, across the three conditions in the study -- CF-Single, CF-Group, and CF-Group-Hint -- significant improvements were observed in people's accuracy as they moved from the training to the testing phases of the experiment, an improvement that trends higher in the two CF-group conditions. However, though significant trends are found in accuracy, the relative differences between the three conditions are modest. This finding may be due in part to ceiling effects; that is, participants' accuracy was initially high in the training phase, limiting the extent to which it could improve from the provision of explanations. Interestingly, accuracy levels for item-groups were found to be strongly correlated with the sequence in which they appeared to participants in the CF-group conditions. The manipulation also showed modest improvements in subjective judgments of confidence, explanation satisfaction and trust, with significant trends found across the three conditions.

\subsection{Objective Measure: Accuracy}
\label{sec:objective}
\begin{figure}[!h]
\centering
\includegraphics[width=0.45\textwidth]{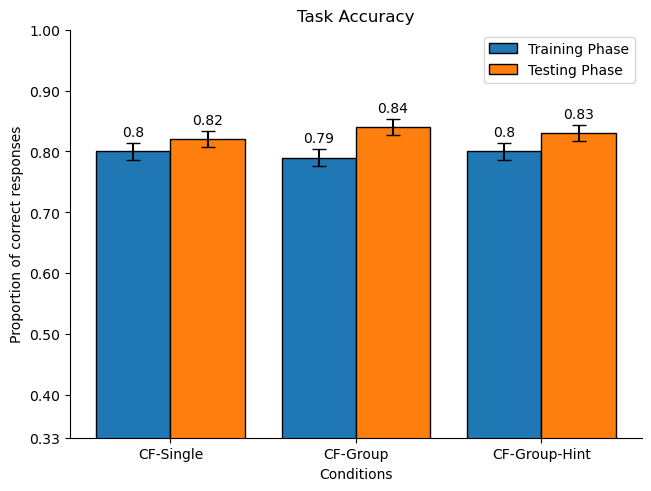}
\centering
\caption{Task Accuracy (proportion of correct answers) for the three conditions (CF-Single, CF-Group, CF-Group-Hint) in the Training and Testing Phases of the Study (error bars show standard error of the mean; y-axis begins at 0.33 to show chance-level for responding).}
\label{fig:fig5}
\Description{x}
\end{figure}

A 3 (Explanation: CF-Single vs CF-Group vs CF-Group-Hint) x 2 (Phase: Training vs Testing) mixed ANOVA with repeated measures on the second factor was carried out on the proportion of correct responses given by each participant (i.e., accuracy; see Figure \ref{fig:fig5}). There was no main effect of Explanation \textit{F}(2, 204)=.174, \textit{p}=.840, however, there was a main effect of Phase, \textit{F}(1, 204)=25.153, \textit{p}<.001, $\eta_p^2$=.11. Participants in all three conditions were more accurate in the testing phase (\textit{M}=.829, \textit{SD}=.096) than in the training phase (\textit{M}=.795, \textit{SD}=.111). The two factors did not interact \textit{F}(2, 204)=1.608, \textit{p}=.203.

Notably, response accuracy was quite high from the outset in the training phase ($\sim$80\% in all three conditions), potentially leading to ceiling effects in the testing phase. The margin for improvement was low, reducing room for differences between the conditions to emerge. However, in the testing phase, there was evidence of a trend in increasing accuracy across conditions with the following order: CF-Single $<$ CF-Group $<$ CF-Group-Hint conditions, Page's L(40)=500.0, \textit{p}=.013 (see Figure \ref{fig:fig5}). This trend indicates that when people were given group counterfactual explanations (without and with a hint), their accuracy progressively improved. Indeed, examining the improvement in accuracy (the difference between a given participant's training accuracy and their testing accuracy), we observe that the CF-Single condition shows the least improvement (\textit{M}=0.019, \textit{SD}=.089), with the CF-Group (\textit{M}=0.049, \textit{SD}=.083) and CF-Group-Hint conditions scoring higher (\textit{M}=0.034, \textit{SD}=.117)).

To tunnel further into these effects, we conducted exploratory analyses on the specific item-sets used in the training phase of the study. In the training phase, participants were presented with eight distinct 5-item-sets (i.e., 40 items in total); that is, sets of 5 similar instances and predictions that were explained with the same group counterfactual in the experimental conditions (CF-Group and CF-Group-Hint) and matched with respect to the control condition (CF-Single). For each participant, the 40 items were randomly re-ordered to control for possible order effects between item-sets. However, due to this randomisation, participants would have seen more and less favourable sequences for a given item-set; that is, the 5 items in a set could happen to be presented together in the randomised sequence (with no gaps between them) while another item-set could be interspersed with instances from other item-sets (with many gaps between items from the same set). Importantly, in the CF-Group and CF-Group-Hint conditions, this would mean that a participant could be presented with five (grouped) counterfactual explanations one after the other using the same target-feature values, presumably making it easier for them to benefit from the group-counterfactual. So, if a given item-set has lower gap-scores, one would expect higher accuracy for that set in both CF-group conditions.

Accordingly, we analysed the order of items presented and calculated  gap-scores for each item-set presented to participants in the three conditions of the study to check whether favourable orderings had any effect. Spearman's correlations were computed between these gap-scores for item-sets in the training phase and the accuracy observed in that phase. This analysis showed that there were negative correlations between gap-scores and accuracy in all three conditions, the lower the gap-score between items from the same set, the higher the accuracy: CF-Single (\textit{r\textsubscript{S}}(6)=-0.43), CF-Group (\textit{r\textsubscript{S}}(6)=-0.38), and CF-Group-Hint (\textit{r\textsubscript{S}}(6)=-0.69). Here, moderate correlations are found in the CF-Single and CF-Group conditions, with a high correlation being found in the CF-Group-Hint conditions, in which participants were told that certain items were part of a group. While this correlation does not imply a causal relationship, it does present a consistent picture of the effects of group counterfactual explanations with a hint. Indeed, this finding suggests that XAI systems should present instances that are counterfactually-grouped in an unbroken sequence, to allow end-users to benefit from this ordering effect.

\subsection{Subjective Measures}
\label{sec:subjective}
A series of subjective evaluations were made by participants in the study, comprising confidence in their responses and satisfaction and trust in the AI system (see Table \ref{tab:mean_scores} for means and standard deviations). Overall, these measures show no main effects between conditions, though significant trends are found in increasing scores across conditions with the following order: CF-Single $<$ CF-Group $<$ CF-Group-Hint.

\textbf{Confidence Judgments} for each prediction in the testing phase were measured using a 5-point Likert scale. A one-way ANOVA conducted on the mean confidence judgments made by participants showed that there was no main effect of group \textit{F}(2, 204)=1.424, \textit{p}=.243. However, a reliable trend was found in the judgments made across conditions in the order: CF-Single $<$ CF-Group $<$ CF-Group-Hint, Page's L(40)=513.5, \textit{p}<.001. Participants showed increasing confidence that their own predictions were correct after they received group-counterfactuals with hints about instances being in a group, with the other conditions showing equivalent scores.

\textbf{Explanation Satisfaction} for the explanations given for the AI system was assessed after the testing phase using the 8-question DARPA scale \cite{hoffman2018metrics}. A one-way ANOVA conducted on the summed scores showed there was no main effect of group \textit{F}(2, 204)=1.424, \textit{p}=.243. However, a reliable trend was found in the judgments made across conditions in the order: CF-Single $<$ CF-Group $<$ CF-Group-Hint, Page's L(8)=105.0, \textit{p}=.012. Here, participants showed increasing satisfaction in the explanations of the AI system, especially when they have received group counterfactuals with hints, with the other two conditions showing equivalent responding.

\textbf{Trust} in the AI system was assessed following the testing phase using the 8-question DARPA scale \cite{hoffman2018metrics}. A one-way ANOVA conducted on the summed scores showed there was no main effect of group \textit{F}(2, 204)= 1.496, \textit{p}=.226. However, again, a reliable trend was found in the judgments made across conditions in the order: CF-Single $<$ CF-Group $<$ CF-Group-Hint, Page's L(8)=108.0, \textit{p}=.001. Here, participants showed increasing trust in the AI system when they received group counterfactuals with hints, with the other two conditions showing equivalent responding.

\section{Related Work}
\label{related_work}
The related work can be split into strands of prior computational and psychological research. In the computational literature \cite{keane2021if,verma2022counterfactual,karimi2022survey, guidotti2022counterfactual}, the traditional use case for counterfactual XAI tries to \textit{explain a single predictive-instance from a single AI model using a single counterfactual} (i.e., using the feature-differences in the counterfactual-pair that change the prediction) with the typical stakeholder being a customer end-user of that AI system (e.g., bank customer). The use case explored in this paper is directed at the same customer stakeholder but is \textit{explaining \textbf{multiple, similar predictive-instances} from a single AI model using a \textbf{single group-counterfactual}} (i.e., using a common feature-value in the difference-features that change the prediction; see Figure \ref{fig:fig1}). As we shall see in the next subsection, there are a few papers that advance related (but different) algorithms addressing other use cases and stakeholder groups (e.g., end-users versus modellers) in the literature. In the  user-study literature, several papers report psychological tests of counterfactual XAI (e.g., \cite{van2021evaluating,kuhl2022keep,Warren2023Features2,Dai2022CounterfactualXAI}) but none of this prior work explicitly tests the type of counterfactuals examined here (i.e., the group-counterfactual idea proposed in focus groups, see section \ref{when_group_exp}). 

\subsection{Computational: Related Algorithms}
Many different methods have been proposed in XAI to explain single instance-prediction from a single AI model using a single counterfactual: methods adopting optimisation approaches \cite{wachter2017counterfactual,mothilal2020explaining,dandl2020multi,karimi2021algorithmiccf}, instance-based approaches \cite{keane2020good, delaney2021instance,nugent_case-based_2005,martens2014explaining,brughmans2021nice}, distributional approaches \cite{dhurandhar2018explanations, joshi2019towards, Poyiadzi2020} and other approaches \cite{guidotti2018local,guidotti2019factual} (for reviews see \cite{keane2021if,verma2022counterfactual,karimi2022survey, guidotti2022counterfactual}).  However, none of this work explains multiple similar, predictive-instances by grouping them using a group-counterfactual, although some prior work has considered related problems for different use cases and stakeholders.

Artelt et al. \cite{artelt2022one} consider using a counterfactual to explain predictions by an ensemble of distinct models; their domain involved a water-distribution system, wherein different sections were modelled in the ensemble, relying on different sensed data, some of which come from faulty sensors. The Ensemble Consistent Explanations (ECEs) method produces a single counterfactual explanation for this set of distinct, individual predictions, a type of summary of the diverse decisions by the different models. The stakeholder in this use case appears to be a model-user deploying the ensemble (i.e., an engineer or statistician). Furthermore, the method attempts to solve a different problem to the current one, as it is \textit{explaining multiple, \textbf{different} predictive-instances from \textbf{multiple} AI models using a single counterfactual} (that is not a group-counterfactual in our sense). It should also be noted that ECEs does not seem to produce very sparse counterfactuals ($>$ 20 feature-differences are common) relative to simpler baselines (that use $\sim$3 feature-differences), so it is unclear whether non-expert end-users would find them easy to understand \cite{keane2020good,forster2021capturing}.

In other related work, Rawal \& Lakkaraju \cite{rawal2020beyond} propose the Actionable Recourse Summaries (AReS) framework to address a different use case for other stakeholders. Their method constructs global counterfactual rules that are interpretable recourse summaries for subgroups in the dataset; here, these subgroups are defined by features-of-interest (e.g., race, gender) with a view to auditing datasets for bias. AReS finds compact feature-difference rule-sets that capture counterfactual recourse for sub-populations of the dataset; for instance, showing that the recourse for foreign worker borrowers (e.g., to earn 20\% more in income) might be less favourable than that for non-foreign workers (e.g., to earn 5\% more in income). This method also assesses fairness using recourse costs based on the actionability of different feature-changes. A small user study (N=21) is reported showing that the recourse summaries generated can aid model developers in detecting model bias and discrimination (though it is unclear whether these differences are statistically significant, or if the study is sufficiently powered to detect them). Here, by comparison to the current work, the stakeholder appears to be quite different as it is a model-developer or AI system auditor. Furthermore, the method again aims to solve a different problem to the current one, as it is \textit{explaining selected \textbf{instance subgroups} from a single AI model using a \textbf{counterfactual summary}} (i.e., abstracted feature-differences expressed as rules), rather than a group-counterfactual. AReS is a very interesting solution to this recourse problem but addresses very different concerns to those handled by the current method.

Plumb et. al. \cite{plumb2020explaining} propose another counterfactual-summary method for a different use case with the same stakeholders (i.e., model developers). Their use case deals with the exploration of groups of points in low-dimensional representations in the ML pipeline. Using an initial model they propose Global Counterfactual Explanations (GCEs) to represent key feature-differences between sub-groups of points in the dataset, which are computed from what they call Transitive Global Translations (TGT). This method computes counterfactual summaries to assess the utility of the low-dimensional representations being used in an initial model. So, as with AReS, the stakeholder appears to be a model developer and the method is again attempting to solve a different problem to the current one; it is \textit{explaining \textbf{subgroups of data-points} from a single AI model using a \textbf{counterfactual summary}}, rather than a group-counterfactual. 

\subsection{Psychological: User Studies}
Although many criticisms have been made of the paucity of user studies in counterfactual XAI \cite{keane2021if}, more well-designed studies have begun to appear in response \cite{Warren2023Features2,Warren2022FeaturesXAI,Dai2022CounterfactualXAI,celar_cfactual23,van2021evaluating,lage2019evaluation,lucic2020does,forster2020evaluating,forster2021capturing,kuhl2022keep,kuhl2022let,le2022improving}. These studies address the traditional counterfactual XAI use case where an AI system's prediction for a single instance is explained using a single counterfactual (e.g., a loan application that is refused, explained with a counterfactual such as "if you had earned more income you would have been granted the loan"). As such, none of this work has considered the use case examined here, where a user is presented with many counterfactual explanations for different predictive-instances in which the same feature-differences are being used over and over counterfactual pairs (see Figure \ref{fig:fig1}). Indeed, in the wider psychological literature, we have not been able to find any work that specifically examines repeated explanations of this sort.

However, there is a strand of explanation research in cognitive science that has argued that explanations that are broad, simple and cover more phenomena are to be preferred \cite{lombrozo2006structure,lombrozo2016explanatory,thagard1989extending,keil2006explanation}. Lombrozo \cite{lombrozo2016explanatory} has argued that a good explanation accounts for a broad range of observations; for example, explanations in the form of disease diagnoses that account for three observed symptoms are judged to be better than those that account for just one \cite{read1993breadth, edwards2019explanation}. Providing a repeated counterfactual explanation based on the same feature-differences, that covers multiple predictive-instances, allows people to form a common explanation that can be applied across similar examples they may encounter. Similarly, group-counterfactuals are simpler by virtue of this repetition too. As such, this literature provides background support for why people might find group counterfactuals useful in explaining the predictions of an AI system, allowing them to learn more about how the domain or system works (the main finding of the current user study).

\section{Conclusions, Caveats \& Future Work}
\label{conclusions}
Based on user feedback, we have advanced a new use case for counterfactual XAI that groups predictive-instances using common counterfactual explanations. A novel algorithm  has been implemented for this use case and applied to generate explanations that were tested in a carefully-controlled user study. This evaluation shows that group-counterfactuals elicit modest but definite improvements in people's understanding of an AI system, as evidenced by people being more (i) accurate in their own predictions, (ii) confident about their answers, and (iii) satisfied with and trusting in the system's performance. Given these results, we outline several conclusions, caveats, and proposals for future work.

First, this work opens up a new frontier in the use of counterfactual explanations in XAI. Interestingly, it highlights the importance of recognising that sometimes users may be given multiple explanations and that it makes sense to group these explanations in some way. In current counterfactual XAI research, the standard scenario adopted is a one-shot, one-way interaction between the AI system and a user; the user receives a prediction and an explanation of it, without accounting for similar instances and their explanations. Beyond this restrictive one-shot paradigm, we have uncovered one real-world scenario in which an AI system is making repeated predictions for queries that can be meaningfully grouped together to provide additional insights to users. In order to provide useful explanation and recourse, we believe that XAI research needs to move beyond traditional, simplified scenarios to more complex real-world, user-driven solutions \cite{doshi2017towards}.

Second, the current implementation, \textbf{Group-CF}, is clearly just the first of many potential algorithmic solutions to this multi-explanation use case. The present solution consciously adopted an instance-based approach, in which users are presented with explanations that specifically modify query instances in the same way, indicating their grouping by using the same counterfactual transformation repeatedly (i.e., substituting the same target-value for multiple predictive-instances). Many other options are possible; for example, the group-counterfactual could be based on showing ranges of values or generalisations of feature-differences computed over sets of related instances. While the core focus of this work was to establish the utility of group counterfactual explanations through a user study and the contribution of a novel technique, additional algorithmic testing (e.g., robustness experiments \cite{slack2021counterfactual}), could further motivate the utility of group explanations beyond single counterfactuals. We would encourage the community to explore these other options to advance the area. 

Third, although the present user study provides firm empirical support for the use of group counterfactuals, it remains to be seen whether stronger effects can be found for other datasets and task contexts.  We posit that the current findings were inhibited by ceiling effects; that is, from the outset people were already quite accurate in predicting outcomes (during the training phase) leaving them very little headroom for improvement. In other domains, where users have less expertise or where the interaction between features is more complex, we would predict stronger effects from supporting group-counterfactual explanations. Furthermore, these group counterfactuals could be presented in different ways; for example, in our tests, these common grouped-explanations typically were not in a sequence together, but if the context were to allow predictive-instances to be grouped in advance to appear together (e.g., sick cows of a similar age or breed), increased benefits of group counterfactuals for users may emerge.

Finally, we would like to join recent calls for a more user-centred XAI \cite{MeiLiptonNorthStar2023,wang2019designing,schoon2021human}. To date, XAI research has been dominated by a ``mad dash'' to implement various explanation strategies; witness the $\sim$120 distinct methods in the counterfactual XAI literature. It is entirely possible that many of these implementations are indistinguishable from a psychological perspective; that is, many of these algorithmic variants would not change people's responses in any way. As explanations must be ultimately understood by their intended users, we emphasise the importance of appropriate user evaluation when designing new explanation methods (see e.g., \cite{doshi2017towards, Liao_Zhang_Luss_Doshi-Velez_Dhurandhar_2022}). One of the take-home messages from the current work is that even a small amount of work on soliciting user views can lead to significant advances in use-case definition and algorithmic development.

\section{Acknowledgements}
This publication has emanated from research conducted
with the financial support of (i) the UCD Foundation, (ii) Science Foundation Ireland
(SFI) to the Insight Centre for Data Analytics under Grant
Number 12/RC/2289 P2 and (iii) SFI and the Department of
Agriculture, Food and Marine on behalf of the Government
of Ireland under Grant Number 16/RC/3835 (VistaMilk).
%
%
\bibliographystyle{ACM-Reference-Format}
\bibliography{paper_draft}

\end{document}